\documentclass[journal,twoside,web]{ieeecolor}
\usepackage{generic}
\usepackage{cite}

\usepackage{amsmath,amssymb,amsfonts}
\usepackage{algorithmic}
\usepackage{graphicx}
\usepackage{textcomp}

\usepackage{newfloat}
\usepackage{listings}
\usepackage{amsmath}
\usepackage{dsfont}
\usepackage{multirow}

\def\BibTeX{{\rm B\kern-.05em{\sc i\kern-.025em b}\kern-.08em
    T\kern-.1667em\lower.7ex\hbox{E}\kern-.125emX}}
\begin{document}
\title{Parametric Primitive Analysis of CAD Sketches with Vision Transformer}
\author{Xiaogang Wang, Liang Wang, Hongyu Wu, Guoqiang Xiao and Kai Xu
\thanks{Manuscript received 05 January 2024; revised 26 March 2024; accepted 24 May 2024. This work was supported in part by the NSFC (62102328, 62325211, 62132021), in part by the Fundamental Research Funds for the Central Universities (No. SWU120076), 
in part by the Open Project Program of State Key Laboratory of Virtual Reality Technology and Systems, Beihang University (No.VRLAB2023C01) and the Major Program of Xiangjiang Laboratory (23XJ01009). Paper no. TII-24-0066. (Corresponding author: Xiaogang Wang and Kai Xu.)}
\thanks{Xiaogang Wang, Liang Wang and Guoqiang Xiao are with the College of Computer and Information Science, Southwest University, Chongqing 400715, China (wangxiaogang@swu.edu.cn, wliang@email.swu.edu.cn, gqxiao@swu.edu.cn)}
\thanks{Hongyu Wu is with the State Key Laboratory of Virtual Reality Technology and Systems, Beihang University, Beijing 100191, China (e-mail: whyvrlab@buaa.edu.cn).}
\thanks{Kai Xu is with the National University of Defense Technology, Changsha 410073, China (e-mail: kevin.kai.xu@gmail.com).}}

\maketitle
\begin{abstract}
The design and analysis of Computer-Aided Design (CAD) sketches play a crucial role in industrial product design, primarily involving CAD primitives and their inter-primitive constraints. To address challenges related to error accumulation in autoregressive models and the complexities associated with self-supervised model design for this task, we propose a two-stage network framework. This framework consists of a primitive network and a constraint network, transforming the sketch analysis task into a set prediction problem to enhance the effective handling of primitives and constraints. By decoupling target types from parameters, the model gains increased flexibility and optimization while reducing complexity. Additionally, the constraint network incorporates a pointer module to explicitly indicate the relationship between constraint parameters and primitive indices, enhancing interpretability and performance. Qualitative and quantitative analyses on two publicly available datasets demonstrate the superiority of this method. 
\end{abstract}


\section{Introduction}
\label{sec:introduction}
\IEEEPARstart{A}{S} mentioned by \cite{yang2022paradigm}, ``The efficient mining of high-value information within industrial Big Data and the utilization of ground truth-life industrial processes are currently some of the most trending topics.'' In this context, the introduction of a large volume of CAD sketch data into the industrial domain will bring about significant opportunities and challenges. In industrial design, CAD sketches are widely utilized for product design and development. However, existing CAD design software demands extensive professional knowledge, involving a series of complex interactions, posing significant barriers and challenges for ordinary users to participate in the design process. Hand-drawn sketches can easily complete designs without the need for professional software or extensive professional knowledge. However, a crucial prerequisite is quickly converting hand-drawn sketches into high-quality CAD drawings. In this work, we propose a neural framework that can automatically parse the CAD primitives and their constraints directly from hand-drawn sketches.

In addition to primitives, CAD sketches also include constraint relationships between primitives, such as Coincident, Parallel, and other constraints (as shown in Table \ref{tab:table_Four}). Subsequently, a series of operations such as extrusion and trimming are applied to 2D sketches to convert them into 3D CAD models gradually. While the current paradigm for CAD sketching is undeniably powerful, it is often a challenging and tedious process. Engineers typically initiate visualization with rough hand-drawn sketches. Therefore, automatically and reliably converting such hand-drawn sketches into parameterized, editable CAD sketches is a highly useful capability.

With the development of deep learning technology, solving the above task has become possible; meanwhile, a high-quality CAD sketch dataset is also a necessary condition. Recently, Vitruvion \cite{seff2021vitruvion} constructed a large-scale dataset, SketchGraph \cite{seff2020sketchgraphs}, by extracting a large number of CAD sketches from Onshape, laying the foundation for sketch analysis tasks. Additionally, Vitruvion \cite{seff2021vitruvion} introduces a generative modeling approach for parameterized CAD sketch generation, utilizing an autoregressive network to construct primitive and constraint models. During the training/inference phase, Vitruvion \cite{seff2021vitruvion} requires the results of the previous $N$-1 steps as input to infer the result of the N-$th$ step, repeating this process to predict the N-$th$ primitive parameters until an end token is encountered. However, this architecture is not only time-consuming but also prone to significant error accumulation if the outputs of earlier steps are biased or incorrect. SketchConcept \cite{yang2022discovering} proposed a complex self-supervised network framework for generating primitives relation graph for CAD sketch analysis. However, due to the complexity of this network architecture design, training is very unstable, and the network output is not very intuitive, making it difficult to directly apply to hand-drawn sketch analysis tasks.

\begin{figure*}[h]
    \centering
    \includegraphics[width=\textwidth,keepaspectratio]{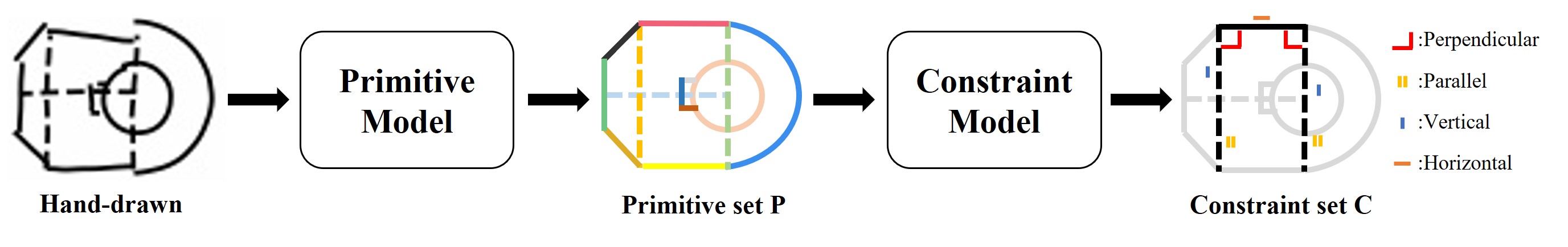}
    \caption{The system framework of our approach. The primitive model takes a hand-drawn sketch as input to produce a parameterized primitive set $P$. Then, the constraint model takes the primitive set $P$ as input to predict the constraint set $C$ of the primitives.}
    \label{fig:image_flow_chart}
    \vspace*{\fill}
\end{figure*}

We reformulate the task of hand-drawn sketch analysis as a set prediction problem. This allows for end-to-end parallel output of all primitives and constraints in the sketch, ensuring that each output result is independent and does not interfere with others. Additionally, we introduce a pointer module into the network to establish the relationship between the predicted results and the primitive indices. This module enables the direct prediction of constraint parameters (primitive indices), indicating which indices correspond to primitives that have constraint relationships. Our approach not only avoids the iterative complexity and error accumulation issues faced by Vitruvion \cite{seff2021vitruvion}, but also circumvents the complexity in design and use seen in SketchConcept \cite{yang2022discovering}.

The representation of primitives is the prerequisite for the design of network algorithms. In this work, the primitive representation mainly includes: type, purpose, and parameters, as shown in Table \ref{tab:table_one}. The purpose of a primitive mainly refers to whether the given primitive represents a physical entity or serves as a reference for constraints. The parameters of a primitive mainly refer to its coordinates, such as the coordinates of the two endpoints of a line primitive. In industrial design, the most common primitive types are line, circle, arc, and point. In addition to these types, the free-form curve is also a crucial type. However, existing references and datasets (e.g., Vitruvion \cite{seff2021vitruvion}) focus on four primitive types (line, circle, arc, and point). To ensure a fair comparison with public algorithms, we maintained experimental consistency with previous works in this study.

However, our primitive representation method can also fully represent free-form curve types. Specifically, unlike other primitive types, the number of parameters for free-form curves is unknown (the number of control points or value points). Therefore, we can treat free-form curves as a prediction problem of an indefinite-length parameter sequence [$x_1$, $y_1$, $x_2$, $y_2$, ..., $x_n$, $y_n$], which is very common in natural language processing (NLP) tasks. We can also use similar approaches for processing. Compared to other primitive types, free-form curve primitives have two differences: 1. the sequence length is indefinite (a maximum parameter number is usually set for ease of processing); 2. similar to NLP tasks, an end symbol $\langle End \rangle$ is added, which effectively controls the end position of the parameter sequence. 
However, existing references and datasets (e.g., Vitruvion \cite{seff2021vitruvion}) focus on four primitive types (line, circle, arc, and point).

The complexity of the sketch analysis task, which not only needs to identify the primitives but also needs to analyze the constraint relationship between the primitives, such as Coincident, Vertical, etc. To this end, we split the complex task into two sub-tasks that are easy to design and complete: 1) Primitive parsing, and 2) Primitive constraint analysis. As shown in Fig. \ref{fig:image_flow_chart}, we are able to extract primitives from rasterized hand-drawn sketches and then further input these extracted primitives into the primitive constraint model, getting the constraint relationships between these primitives. 

The main contributions are summarized as follows:
    \begin{enumerate}
    \item We transform the sketch analysis task into a Set Prediction problem, effectively simplifying the network framework for primitive detection and constraint estimation in sketch analysis.
    \item We decouple the predicted primitive types and parameters in the primitive network, ensuring that they are independent of each other.
    \item We apply a pointer module to our constraint network, associating constraint parameters with the indices of the primitives they affect. It enhances the interpretability of our model while effectively improving performance. 
    \end{enumerate}

In the rest of the paper, we first conduct a review of the previous studies on CAD sketch analysis in Section \ref{Related Work}. In Section \ref{Methodology}, we present the technical details of the proposed networks framework. Experimental results are shown in Section \ref{Experiments}. At last, we conclude with a discussion in Section \ref{Conclusion}.

\section{Related Work}
\label{Related Work}
\subsection{Parametric primitive inference}
The problem of fitting parameterized primitives has been a longstanding challenge in geometric processing. 
PIE-Net \cite{wang2020pie} infers 3D parameterized curves by analyzing point clouds of edges. NEF \cite{ye2023nef} and NerVE \cite{zhu2023nerve} share a common goal with PIE-Net \cite{wang2020pie}, but they differ slightly in their implementation. NEF \cite{ye2023nef} relies on a set of calibrated multi-view images to reconstruct the 3D feature curves of objects. On the other hand, NerVE \cite{zhu2023nerve} utilizes neural volume edges to map point clouds into a unified free-form curve, thereby obtaining the final target parameter curves. ParSeNet \cite{sharma2020parsenet} decomposes 3D point clouds into sets of parametric surface patches. Point2cly \cite{uy2022point2cyl} segments point clouds into patches corresponding to different primitives, conducting primitive type classification and parameter regression for each patch. SED-Net \cite{li2023surface} incorporates a two-stage feature fusion mechanism, integrating per-point features and edge types into the instance segmentation branch and fitting the segmentation results with parameterized primitives. ComplexGen \cite{guo2022complexgen} adopts a boundary representation (B-Rep) strategy to simultaneously recover corner points, curves, and surfaces along with their topological constraints. These approaches share the common feature of utilizing 3D point cloud data as input. Our goal aligns with that of Vitruvion \cite{seff2021vitruvion}, which is to extract parameterized primitives from ground truth hand-drawn sketches.

\subsection{Base sketch}
ZS-SBPRnet \cite{peng2022zs} proposes a novel zero-shot sketch-based point cloud retrieval network based on feature projection and cross reconstruction. SECAD-Net \cite{li2023secad} forms the final 3D CAD solid by extruding the 2D plane where the sketch is located. It interpolates the latent space where the sketch is located, enabling the creation of any CAD variant. \cite{willis2021engineering} employing sketch generation models, it becomes possible to transform 2D sketch primitives into 3D entities and ultimately compose a final 3D model by combining multiple entities. A series of recent works \cite{ganin2021computer}, \cite{para2021sketchgen}, \cite{seff2021vitruvion} use autoregressive models \cite{vaswani2017attention} to generate CAD sketches and constraints modeled through pointer networks \cite{vinyals2015pointer}. However, these methods based on autoregressive models require post-processing of the generated results, namely segmenting the network outputs to ensure the independence of each type and its corresponding parameters in the generated sketch. In contrast, our approach is capable of obtaining end-to-end output results, thus avoiding the potential additional errors introduced by network post-processing.

\subsection{DETR-based models}
Recently, DETR \cite{carion2020end} introduced the Transformer \cite{vaswani2017attention} architecture into the field of object detection, ingeniously creating a concise and innovative end-to-end framework that successfully eliminates the laborious manual processes of anchor generation and post-processing steps. Subsequently, numerous researchers have delved into DETR \cite{carion2020end}, aiming to enhance its training convergence and performance further. For instance \cite{wang2022anchor, zhu2020deformable, liu2022dab, li2022dn, zhang2022dino} have respectively made improvements to DETR \cite{carion2020end} in various aspects. Among these, DN-DETR \cite{li2022dn} employs a clever approach by inputting noisy ground truth (GT) bounding boxes into the network and training it to reconstruct the original boxes, effectively reducing the complexity of the binary matching process. These methods primarily focus on the COCO dataset for sequentially detecting different objects. Furthermore, some studies apply the principles of DETR \cite{carion2020end} to other domains. For example, \cite{giard2023electric} identifies the type of fuse through its physical characteristics. SketchConcept \cite{yang2022discovering} extends the DETR \cite{carion2020end} concept to raw sketches, achieving the detection of modular CAD sketch concepts through self-supervised learning. In this context, we propose a method to extend the DETR \cite{carion2020end} architecture further, with focusing on end-to-end detection of primitives or constraint relationships between primitives from a given input.

\begin{figure}[t]
    \centering
    \includegraphics[width=0.49\textwidth,keepaspectratio]{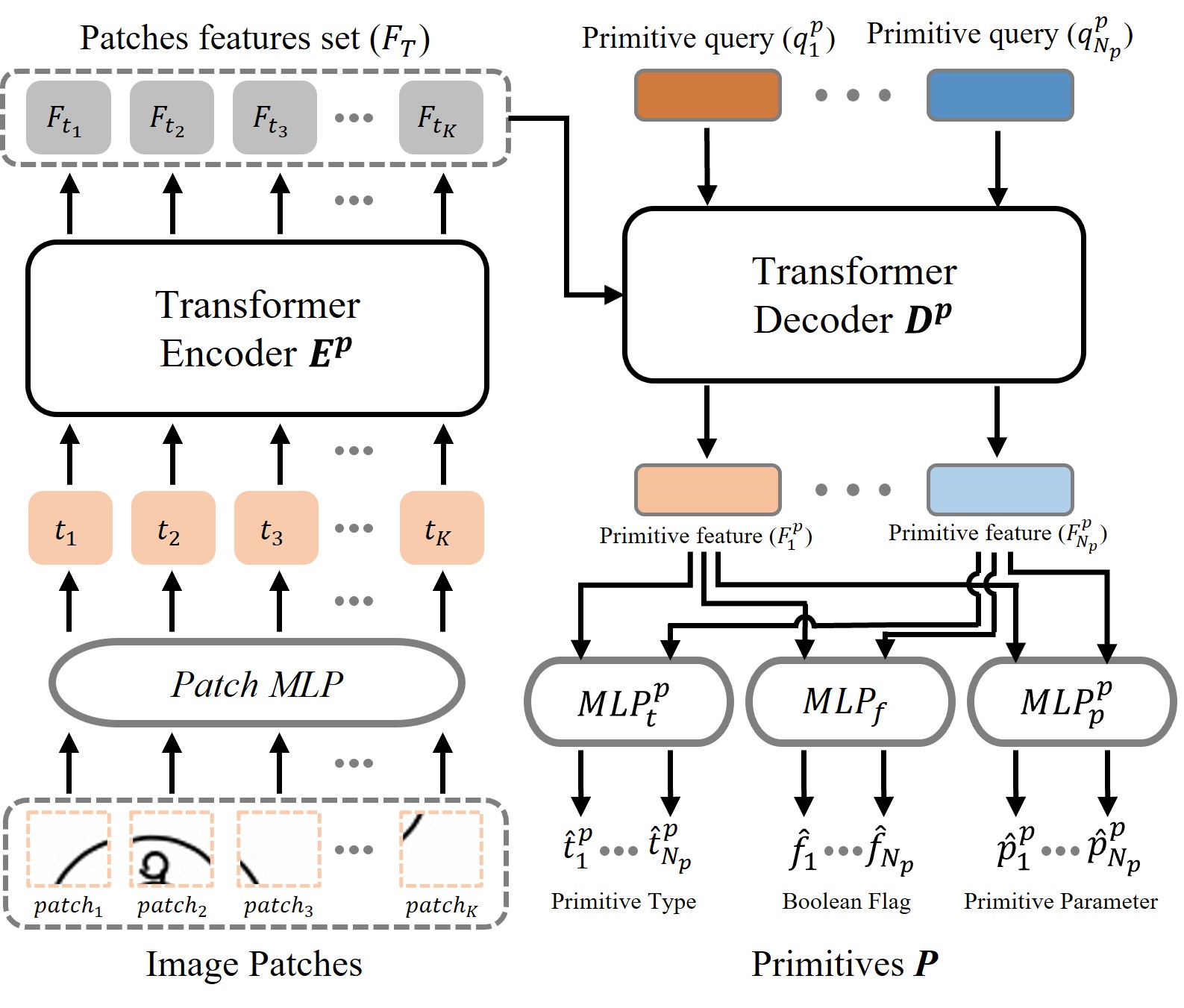}
    \caption{The network architecture of primitive model. The primitive model takes image patches of hand-drawn sketch as input to extract the primitives set $P$.} 
    \label{fig:primitive_pipeline}
\end{figure}
\section{Methodology}
\label{Methodology}
Each sketch, denoted as $S$, consists of both a set of primitives, denoted as $P$, and a set of constraints, denoted as $C$. Noteworthy, there is no explicit order among primitives, as different designers often have different design habits for the same object. We define primitives set $P$ and constraints set $C$ using the following expressions: $P = \{P_i\}_{i=1}^{K_p}$ and $C = \{C_i\}_{i=1}^{K_c}$, $K_p$ and $K_c$ represent the actual number of primitives and constraints contained in the sketch $S$, respectively. As shown in Table \ref{tab:table_one}, each primitive $P_j=(t^{p}_j,f_j,p^{p}_j)$ specifies the primitive type $t^{p}_j$, its boolean flag $f_j$ (Boolean specifies whether the given primitive represents a physical entity or served as a reference for constraints.) and its primitive parameters $p^{p}_j$. As shown in Table \ref{tab:table_Four}, each constraint $C_j=(t^{c}_j,p^{c}_j)$ specifies the constraint type $t^{c}_j$ and its constraint parameters $p^{c}_j$. Primitives and constraints are represented similarly in that both have a definite type (e.g., Line, Arc, Coincident, Horizontal) as well as a set of additional parameters that determine their placement/behavior in a sketch.

We divide the sketch analysis task into two subtasks: Primitive parsing, and Primitive constraint analysis. Their setup is similar to DETR \cite{carion2020end}, which detects objects from given inputs.
\begin{table}[t]
    \centering
    \caption{Primitive parameterization table.}
    \resizebox{\linewidth}{!}{
    \begin{tabular}{c|c|c|c|c|c|c|c|c}
    \hline
        \textbf{Type} & \textbf{Boolean} & \multicolumn{7}{c}{\textbf{Parameters}} \\ 
        \hline
        Line & $flag$& $x_1$ & $y_1$ & $x_2$ & $y_2$ & $0$ & $0$& $0$\\ \hline
        Circle & $flag$ & $x_1$   & $y_1$   & $0$   & $0$   & $0$ & $0$& $r$\\ \hline
        Arc    & $flag$& $x_1$ & $y_1$ & $x_{2}$ & $y_{2}$ & $x_3$ & $y_3$& $0$\\ \hline
        Point  & $flag$ & $x_1$   & $y_1$   & $0$   & $0$   & $0$& $0$  & $0$\\ \hline
    \end{tabular}
    }
    \label{tab:table_one}
\end{table}

\subsection{Primitive Model}
The primitive model is designed to extract parameterized primitives information, such as lines, circles, arcs, and points, from the given hand-drawn sketch image. 

\textbf{Parameterization.} We parameterize each primitive in the form outlined in Table \ref{tab:table_one}, primitive type, boolean flag, and primitive parameters. In the Table \ref{tab:table_one}, the parameters for the line primitive consist of two endpoints $(x_1, y_1)$ and $(x_2, y_2)$; for the circle primitive, it includes the center $(x_1, y_1)$ and the radius $r$; for the arc primitive, the parameters are three points $(x_1, y_1)$, $(x_{2}, y_{2})$, and $(x_3, y_3)$; while the point primitive is determined solely by a 2D coordinate $(x_1, y_1)$. In the table, $0$ represents padding.

\textbf{Quantization.} The primitive parameters are quantized into 6-bit integers, and the boolean flags are encoded as 1 (true) or 0 (false). 


\textbf{Architecture.} As shown in Fig. \ref{fig:primitive_pipeline}, the sketch image with a resolution of 128×128 is divided into $K$ non-overlapping square patches, denoted as $Patch = \{patch_i\}_{i=1}^{K}$. These patches are flattened and mapped into a token set $T = \{t_i\}_{i=1}^{K}$ via a Patch Multi-Layer Perceptron (MLP). Subsequently, the token set $T$ is further integrated into a patches features set $F_T = \{F_{t_i}\}_{i=1}^{K}$ by the encoder, which is then fed into the decoder. In the decoder, trainable parameters are simultaneously input as a primitive query set \( Q^p = \{q_i^p\}_{i=1}^{N_p} \). The primitive query set \( Q^p \) is continuously updated during the model training process, ensuring that each primitive query \( q_i^p \) corresponds to a parameterized primitive target in the hand-drawn sketch. Through the decoder, each primitive query \( q_i^p \) is transformed into a primitive feature \( F_i^p \), thereby obtaining the primitive features set \( F^P = \{F_i^p\}_{i=1}^{N_p} \). Finally, the primitive features set \( F^P \) is separately fed into three MLPs (\( MLP_t^p \) predicts the primitive type \( \hat{t}^p \), \( MLP_f \) predicts the boolean flag \( \hat{f} \), and \( MLP_p^p \) predicts the primitive parameters \( \hat{p}^p \)) to predict the primitive set \( \hat{P} = \{\hat{P}_i\}_{i=1}^{N_p} \). Each predicted primitive \( \hat{P}_i = (\hat{t}_i^p, \hat{f}_i, \hat{p}_i^p) \) consists of the primitive type \( \hat{t}_i^p \), boolean flag \( \hat{f}_i \), and primitive parameters \( \hat{p}_i^p \). \( N_p \) represents the upper limit of the predicted number of primitive set by the primitive model, which is significantly greater than the actual number of primitives \( K_p \) contained in the sketch.

\textbf{Patch Encoder.} The encoder \(E^p\) consists of 12 Transformer blocks, each with 8 attention heads and a feedforward dimension of 512. The encoder takes a token set \(T = \{t_i\}_{i=1}^K\) as input and outputs a patches features set \(F_T = \{F_{(t_i)}\}_{i=1}^K\), where each patch feature \(F_{(t_i)}\) has a dimension of \(d_E\).

\begin{figure*}[t]
    \centering
    \includegraphics[width=0.99\textwidth,keepaspectratio]{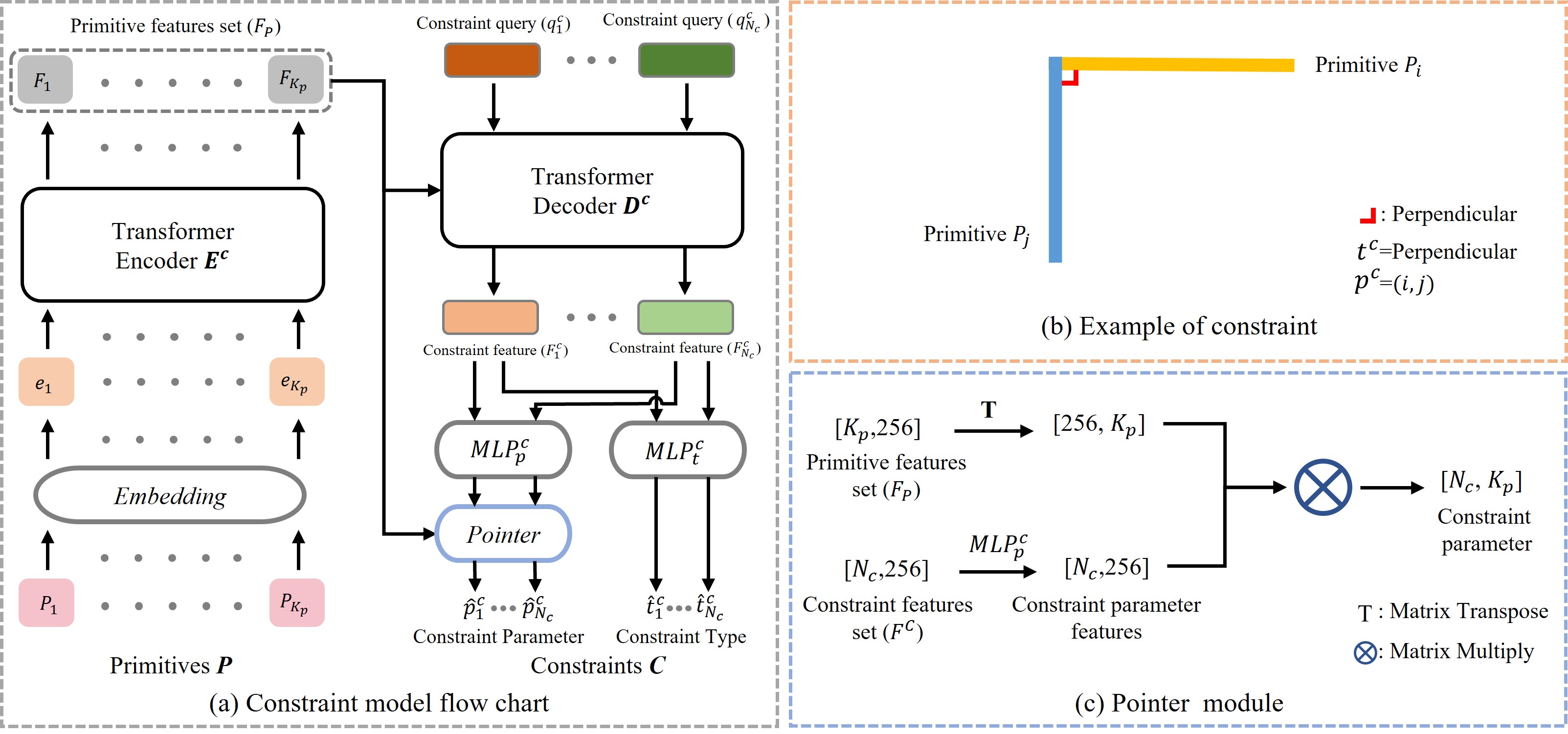}
    \caption{The network architecture of constraint model. The constraint model takes a primitives set $P$ as input to extract the constraints set $C$.} 
    \label{fig:constraint_pipeline}
\end{figure*}

\textbf{Primitive Decoder.} 
We convert the problem of extracting parametric primitive from hand-drawn sketches into a prediction problem of the primitive set. Our decoder $D^p$ has the same hyperparameter settings as the encoder $E^p$. The decoder takes the patches features set \(F_T = \{F_{(t_i)}\}_{i=1}^K\) as input, along with a learnable primitive query set \(Q^p = \{q_i^p\}_{i=1}^{N_p}\). The output primitive features set \(F^P = \{F_i^p\}_{i=1}^{N_p}\) from the decoder is fed into three MLPs separately to predict the primitive set \(\hat{P} = \{\hat{P}_i\}_{i=1}^{N_p}\). Each predicted primitive \(\hat{P}_i\) includes the primitive type \(\hat{t}_i^p\), boolean flag \(\hat{f}_i\), and primitive parameters \(\hat{p}_i^p\), represented by the following equations:
\begin{equation}
        F^P =D^p(F_T,Q^p,Pos_e,Pos_d),
\end{equation}
\begin{equation}
        \hat{t}^p =MLP_t^p(F^P),
\end{equation}
\begin{equation}
        \hat{f} =MLP_f(F^P),
\end{equation}
\begin{equation}
        \hat{p}^p =MLP_p^p(F^P),
\end{equation}
where \(F^P\) represents the primitive features set, \(F_T\) represents the patches features set, \(Q^p\) represents the primitive query set, \(Pos_e\) represents the positional encoding of the patches features set \(F_T\), \(Pos_d\) represents the positional encoding of the primitive query set \(Q^p\), \(\hat{t}^p\) represents the predicted primitive types, \(\hat{f}\) represents the predicted boolean flags, and \(\hat{p}^p\) represents the predicted primitive parameters.

\textbf{Primitive Cost Matrix.} Due to the lack of explicit correspondence between the ground truth (GT) primitive set \({P=\{P_i\}}_{i=1}^{K_p}\) and the predicted primitive set \(\hat{P}=\{\hat{P}_i\}_{i=1}^{N_p}\), which is not possible to compute the loss function directly. To measure the difference between the GT primitive set and the predicted primitive set, we construct a cost matrix between the two sets and represent it using the following formula:
\begin{equation}
        Cost_p =\sum_{i=1}^{K_p}\sum_{j=1}^{N_p}L_{match}(P_i,\hat{P}_j),
        \label{sigma_p}
\end{equation}
where \( K_p \) is the number of primitives contained in the GT sketch, \(N_p\) is the maximum number of predicted primitives set by the primitive model, and \( N_p \) is much larger than \( K_p \). \( L_{match}(P_i, \hat{P}_j) \) is the cost function between the GT primitive \( P_i \) and the predicted primitive \( \hat{P}_j \), obtaining the cost of matching between them. The cost matrix \( Cost_p \) is solved by the Hungarian matching algorithm to obtain the primitive index mapping relationship \( \sigma_p \), where the GT primitive \( P_i \) with index \( i \) corresponds to the predicted primitive \( \hat{P}_{\sigma_p(i)} \) with index \( \sigma_p(i) \) $\in$ \( N_p \).

The cost function \( L_{match}(P_i, \hat{P}_j) \) takes into account the losses of primitive type, boolean flag, and primitive parameters. Each GT primitive is denoted as \( P_i = (t_i^p, f_i, p_i^p) \), where \( t_i^p \) represents the primitive type, \( f_i \) represents the boolean flag, and \( p_i^p \) represents the primitive parameters. For the predicted primitive \( \hat{P}_j = (\hat{t}_j^p, \hat{f}_j, \hat{p}_j^p) \), \( \hat{t}_j^p \) denotes the predicted primitive type, \( \hat{f}_j \) represents the predicted boolean flag, and \(\hat{p}_j^p\) denotes the predicted primitive parameters. The cost function \( L_{match}(P_i, \hat{P}_j) \) is represented by the following formula:
 \begin{align}
L_{match}(P_i,\hat{P}_j) &= \omega^p_{t} * CE(t_i^p,\hat{t}_j^p)  + \omega_{f} * CE(f_i,\hat{f}_j) \nonumber \\
&\quad + \omega^p_{p} * CE(p_i^p,\hat{p}_j^p),
\end{align}
where $CE$ is the standard cross-entropy function, and \( w_* \) is the weight balancing term.

\textbf{Loss.} Through  Eq. \ref{sigma_p}, we obtain the index mapping relationship \( \sigma_p \). Utilizing the index mapping relationship \( \sigma_p \), we correspond the GT primitive \( P_i \) with index \( i \) to the predicted primitive \( \hat{P}_{\sigma_p(i)} \) with index \( \sigma_p(i) \). Based on the mapping relationship between the GT primitive set \( P = \{P_i\}_{i=1}^{K_p} \) and the predicted primitive set \( \hat{P} = \{\hat{P}_i\}_{i=1}^{N_p} \), we compute the primitive type loss \( Loss_{t}^p \), boolean flag loss \( Loss_{f} \), and primitive parameter loss \( Loss_{p}^p \) to calculate the loss function between these two sets. We express as follows:
\begin{equation}
        Loss(P,\hat{P}) = \omega^p_{t}*Loss^p_{t}+\omega_{f}*Loss_{f}+\omega^p_{p}*Loss^p_{p},
        \label{loss_image_primitive}
\end{equation} 
\begin{equation}
        Loss^p_{t} = \sum_{i=1}^{K_{p}}CE(t_i^p,\hat{t}_{\sigma_p(i)}^p),
\label{L^p_{t}}
\end{equation}
\begin{equation}
        Loss_{f} = \sum_{i=1}^{K_p}CE(f_i,\hat{f}_{\sigma_p(i)}),
\end{equation}
\begin{equation}
        Loss^p_{p} = \sum_{i=1}^{K_p}CE(p_i^p,\hat{p}_{\sigma_p(i)}^p),
\end{equation}
where \( \sigma_p \) represents the index mapping relationship, \( t_i^p \) denotes the GT primitive type, \( f_i \) denotes the GT boolean flag, \( p_i^p \) represents the GT primitive parameter, \( \hat{t}_{\sigma_p(i)}^p \) signifies the predicted primitive type, \( \hat{f}_{\sigma_p(i)}^p \) indicates the predicted boolean flag, \( \hat{p}_{\sigma_p(i)}^p \) represents the predicted primitive parameters, \( CE \) denotes the standard cross-entropy function, and \( w_* \) is the weight balancing term.

\begin{table}[t]
\centering
\caption{Constraint parameterization table.} 
\label{tab:table_Four} 
    \begin{tabular}{c|c} 
    \hline
    Type & Parameters \\ \hline
    Coincident  &$(\lambda_1,\lambda_2)$\\ \hline
    Horizontal &$(\lambda_1)$\\ \hline
    Parallel    &$(\lambda_1 ,\lambda_2)$\\ \hline
    ...    &...\\ \hline
    \end{tabular}   
\end{table}
\subsection{Constraint Model}
Only given a primitives set, the constraint model's task is to analyze the constraints between the primitives. 

\textbf{Parameterization.} As shown in Table \ref{tab:table_Four}, we parameterize the constraint between primitives. Each constraint type corresponds to one or two parameters, which are closely related to the primitive index $\lambda_*$. 

\textbf{Architecture.} As illustrated in Fig. \ref{fig:constraint_pipeline} $(a)$, we provide a detailed description of the encoder-decoder Transformer architecture adopted by the constraint model. The constraint model takes the primitive set \( P = \{P_i\}_{i=1}^{K_p} \) as input, where \( K_p \) is the actual number of primitives in the sketch, and each primitive \( P_i = (t_i^p, f_i, p_i^p) \) includes the primitive type \( t_i^p \), boolean flag \( f_i \), and primitive parameters \( p_i^p \) (as shown in Table \ref{tab:table_one}). Initially, the primitive set \( P \) is embedded to form a primitive embedding set \( E = \{e_i\}_{i=1}^{K_p} \), which is then fed into the encoder. The encoder utilizes Transformer to further transform the primitive embedding set \( E \) into a primitive features set \( F_P = \{F_i\}_{i=1}^{K_p} \), serving as input to the decoder. In the decoder, a learnable constraint query set \( Q^c = \{q_i^c\}_{i=1}^{N_c} \) is simultaneously inputted. The constraint query set \( Q^c \) is continuously updated during the model training process, ensuring that each constraint query \( q_i^c \) corresponds to different constraints within the primitive set. Through the decoder, each constraint query \( q_i^c \) is transformed into a constraint feature \( F_i^c \), thus obtaining the constraint features set \( F^C = \{F_i^c\}_{i=1}^{N_c} \). Finally, the constraint features set \( F^C \) is passed through \( MLP_t^c \) and \( MLP_p^c \) to respectively predict the constraint type \( \hat{t}^c \) and the corresponding constraint parameters \( \hat{p}^c \).

As shown in Fig. \ref{fig:constraint_pipeline} $(b)$, we provide an example of constraint. Assuming there exists a Perpendicular constraint between primitive \( P_i \) and \( P_j \), with constraint type \( t^c = Perpendicular \) and constraint parameters \( p^c = (i, j) \), where \( i \) and \( j \) represent primitive indices. Since the constraint parameters are derived from primitive indices, a pointer module \cite{vinyals2015pointer} is employed in Fig. \ref{fig:constraint_pipeline} $(a)$ to emphasize the source of constraint parameters, and the detailed computation process of the pointer module is demonstrated in Fig. \ref{fig:constraint_pipeline} $(c)$.


\textbf{Embeddings.} The constraint model takes the primitives set \( P = \{P_i\}_{i=1}^{K_p} \) as the input. We project each primitive \( P_i \) onto a common embedding space as input to the encoder. Each primitive in the sketch can be represented as \( P_i = (t_i^p, f_i, p_i^p) \), where \( t_i^p \) denotes the primitive type \( P_i \), \( f_i \) represents the boolean flag, and \( p_i^p \) indicates the 7 position-related parameters of the primitive (as shown in Table \ref{tab:table_one}). Each parameter in primitive \( P_i \) can be embedded as the feature vector, and the dimension of the feature vector is \( d_E \). Additionally, the feature vector is learnable during model training.

The embedding vector for the type \( t_i^p \) and the boolean flag \( f_i \) of primitive \( P_i \) are represented as \( E(t_i^p) \) and \( E(f_i) \) respectively, both with a dimension of \( d_E \). The embedding vectors for the parameters \( p_i^p \) of primitive \( P_i \) are represented as \( E(p_i^p) \), with a dimension of \( (7, d_E) \). Here, 7 denotes the number of primitive parameters \( p_i^p \), and \( d_E \) represents the dimension of the embedding vector for each parameter. The embedding vector \( E(P_i) \) for primitive \( P_i \) is defined as follows:
\begin{equation}
    e_i =E(P_i)= \frac{ Sum[E(t_i^p), E(f_i), E(p_i^p)]}{Num},  
\end{equation}
where \( e_i \) represents the embedding vector of primitive \( P_i \) with a dimension of \( d_E \). $Sum$ denotes the summation operation over the embedding vector dimensions. \( Num = 1 + 1 + |p_i^p| \), where the values respectively denote: one primitive type parameter, one boolean flag parameter, and \( |p_i^p| \) primitive parameters (considering only non-padding parameters).

\textbf{Primitive Encoder.} As shown in Fig. \ref{fig:constraint_pipeline} $(a)$, we take the primitive embedding set \( E = \{e_i\}_{i=1}^{K_p} \) as the input to encoder \( E^c \), and after processing by the encoder, we obtain the primitive features set \( F_P = \{F_i\}_{i=1}^{K_p} \). This process is represented by the following formula:
\begin{equation}
        F_p = E^c(Pos_{e},E),
\end{equation}
where \( E \) represents the primitive embedding set, and \( Pos_e \) denotes the positional encoding of \( E \).
\begin{figure}[t]
    \centering
    \includegraphics[width=0.48\textwidth,keepaspectratio]{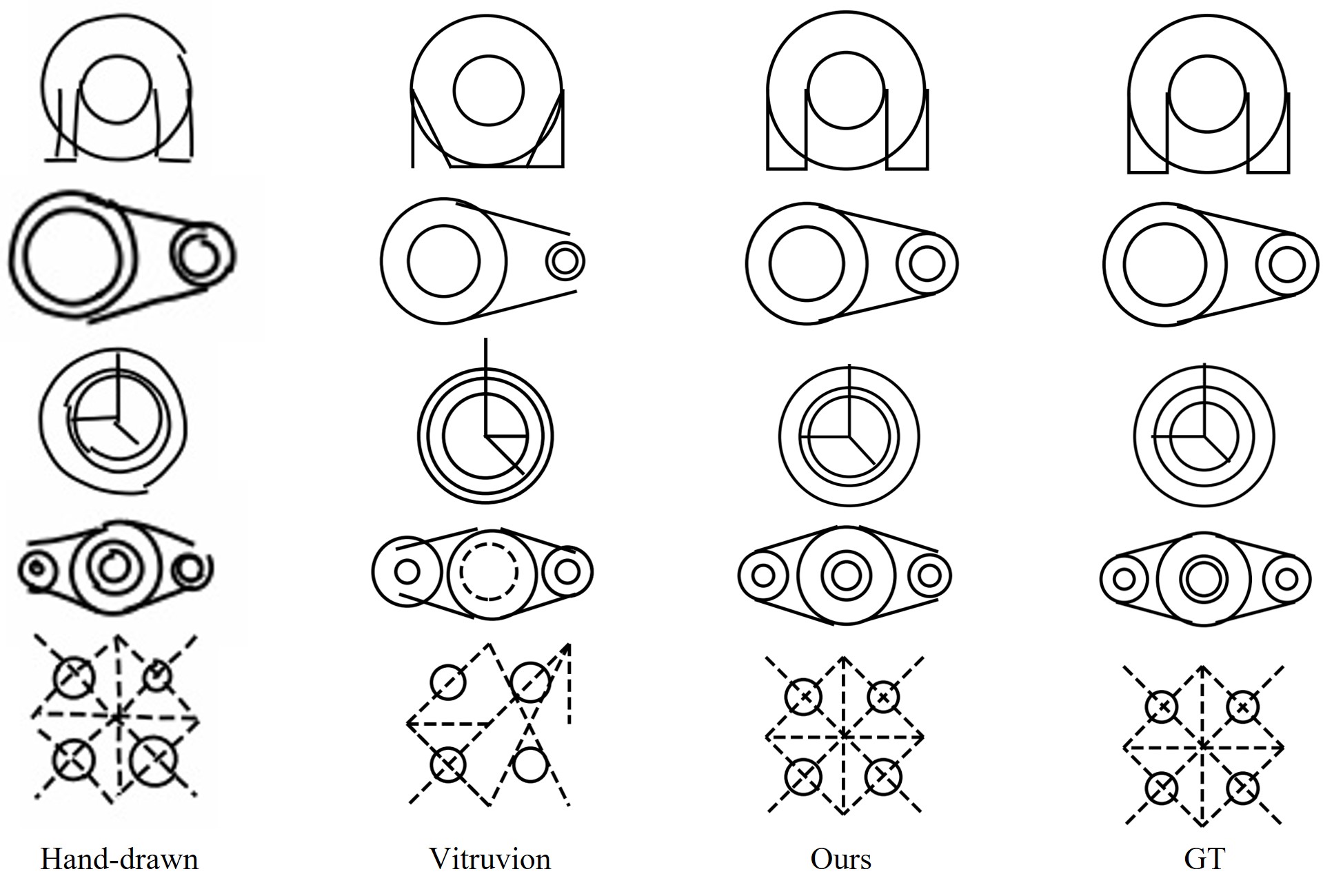}
    \caption{Comparison with a state-of-art primitives parsing method for hand-drawn sketches.}
    \label{fig:primitive_image}
\end{figure}

\begin{table}[t]
    \centering
    \caption{The evaluations of primitive model on the noisy sketch test set of Vitruvion \cite{seff2021vitruvion} and SketchConcept \cite{yang2022discovering} dataset.}
    \resizebox{\linewidth}{!}{
    \begin{tabular}{c|c|c|c|c|c|c}
    \hline
     & \multicolumn{3}{c|}{\begin{tabular}[c]{@{}c@{}}Vitruvion Dataset\end{tabular}} & \multicolumn{3}{c}{\begin{tabular}[c]{@{}c@{}}SketchConcept Dataset\end{tabular}} \\ \hline
    Metrics & $ACC^p_{type}$ & $ACC_{flag}$ & $ACC^p_{par}$ & $ACC^p_{type}$ & $ACC_{flag}$ & $ACC^p_{par}$ \\ \hline
    Vitruvion & 78.75\% & 93.85\% & 49.19\% & 79.36\% & 94.77\% & 47.01\% \\ \hline
    Ours & \textbf{95.11\%} & \textbf{95.00\%} & \textbf{87.26\%} & \textbf{95.45\%} & \textbf{97.48\%} & \textbf{86.76\%} \\ \hline
    \end{tabular}
    \label{tab:table_two_three}
    }
\end{table}

\textbf{Constraints Decoder.} We transform the constraint parsing problem among the primitive set \( P \) into a prediction problem for the constraint set \( C \). As depicted in Fig. \ref{fig:constraint_pipeline} $(a)$, the encoder extracts the primitive features set \( F_P = \{F_i\}_{i=1}^{K_p} \) and feeds it into the decoder \( D^c \). Meanwhile, the decoder receives a learnable constraint query set \( Q^c = \{q_i^c\}_{i=1}^{N_c} \), where each constraint query \( q_i^c \) corresponds to a constraint target in the primitive set, and outputs the constraint features set \( F^C \). This process is represented by the following formula:
\begin{equation}
        F^C = D^c(Q^c,F_P,Pos_{d},Pos_{e}),
\end{equation}
where \( F^C \) represents the constraint features set, \( Q^c \) denotes the constraint query set, \( F_P \) stands for the primitive features set, \( Pos_d \) denotes the positional encoding of the constraint query set \( Q^c \), and \( Pos_e \) represents the positional encoding of the primitive features set \( F_P \).

We use \( MLP_t^c \) to transform the constraint features set \( F^C = \{F_i^c\}_{i=1}^{N_c} \) into the predicted constraint types \( \hat{t}^c \), and utilize \( MLP_p^c \) along with a pointer module \cite{vinyals2015pointer} to convert the constraint features set \( F^C \) into the predicted constraint parameters \( \hat{p}^c \). It is noted that the constraint parameters are derived from primitive indices, where the detailed calculation process of the pointer module is illustrated in Fig. \ref{fig:constraint_pipeline} $(c)$. We express as follows:
\begin{equation}
        \hat{t}^c = MLP^c_t(F^C),
\end{equation}
\begin{equation}
        \hat{p}^c=pointer(MLP^c_p(F^C),F_P),
\end{equation}
where \( F^C \) represents the constraint features set, \( F_P \) denotes the primitive features set, \( \hat{t}^c \) indicates the predicted constraint types, and \( \hat{p}^c \) denotes the predicted constraint parameters.

\textbf{Constraint Cost Matrix.} Due to the lack of explicit correspondence between the ground truth (GT) constraint set \( C = \{C_i\}_{i=1}^{K_c} \) and the predicted constraint set \( \hat{C} = \{\hat{C}_i\}_{i=1}^{N_c} \), it is impossible to compute the loss function. To measure the difference between the GT and the predicted constraint set, we construct a cost matrix between the two sets and represents it using the following formula:
\begin{equation}
        Cost_c =\sum_{i=1}^{K_c}\sum_{j=1}^{N_c}L_{match}(C_i,\hat{C}_j),
        \label{sigma_c}
\end{equation}
where \( K_c \) is the number of constraints contained in the primitive set, \( N_c \) is the maximum number of predicted constraints set by the constraint model, and \( N_c \) is significantly larger than \( K_c \). \( L_{match}(C_i, \hat{C}_j) \) represents the cost function between the GT constraint \( C_i \) and the predicted constraint \( \hat{C}_j \), obtaining the cost value of matching between them. Through the Hungarian matching algorithm to solve the cost matrix \( Cost_c \), we obtain the constraint index mapping relationship \( \sigma_c \), which means the GT constraint \( C_i \) with index \( i \) corresponds to the predicted constraint \( \hat{C}_{\sigma_c(i)} \) with index \( \sigma_c(i) \), where \( \sigma_c(i) \) $\in$ \( N_c \).

The calculation of the cost function \( L_{match}(C_i, \hat{C}_{\sigma_c(i)}) \) needs to consider both the predicted constraint type and the predicted constraint parameters. Each constraint \( C_i \) in the GT constraint set is defined as \( C_i = (t_i^c, p_i^c) \), where \( t_i^c \) represents the GT constraint type and \( p_i^c \) represents the GT constraint parameters. Similarly, each constraint \( \hat{C}_j \) in the predicted constraint set is defined as \( \hat{C}_j = (\hat{t}_j^c, \hat{p}_j^c) \), where \( \hat{t}_j^c \) represents the predicted constraint type and \( \hat{p}_j^c \) represents the predicted constraint parameters. We express as follows:
\begin{equation}
L_{match}(C_i, \hat{C}_j)) = \omega^c_{t}*CE(t_i^c,\hat{t}_j^c) +\omega^c_{p}* CE(p_i^c,\hat{p}_j^c),
\end{equation}
where $CE$ is the standard cross-entropy function, and \( w_* \) is the weight balancing term.

\textbf{Loss.} By utilizing Eq. \ref{sigma_c}, we obtain the index mapping relationship \( \sigma_c \). Through the index mapping relationship \( \sigma_c \), each GT constraint \( C_i \) with index \( i \) is mapped to the predicted constraint \( \hat{C}_{\sigma_c(i)} \) with index \( \sigma_c(i) \). Based on the mapping relationship between the GT constraint set \( C = \{C_i\}_{i=1}^{K_c} \) and the predicted constraint set \( \hat{C} = \{\hat{C}_i\}_{i=1}^{N_c} \), we compute the constraint type loss \( \text{Loss}_t^c \) and the constraint parameter loss \( \text{Loss}_p^c \) separately, facilitating the computation of the loss function between the two sets. We express as follows:
\begin{equation}
        Loss(C,\hat{C}) = \omega^c_{t}*Loss^c_{t}+\omega^c_{p}*Loss^c_{p},
       \label{loss_constraint}
\end{equation} 
\begin{equation}
        Loss^c_{t} = \sum_{i=1}^{K_{c}}CE(t_i^c,\hat{t}_{\sigma_c(i)}^c),
\end{equation}
\begin{equation}
        Loss^c_{p} = \sum_{i=1}^{K_c}CE(p_i^c,\hat{p}_{\sigma_c(i)}^c),
\end{equation}
where \( \sigma_c \) denotes the index mapping relationship, mapping the GT constraint \( C_i \) with index \( i \) to the predicted constraint \( \hat{C}_{\sigma_c(i)} \) with index \( \sigma_c(i) \). \( t_i^c \) represents the GT constraint type, \( p_i^c \) represents the GT constraint parameters, \( \hat{t}_{\sigma_c(i)}^c \) represents the predicted constraint type, \( \hat{p}_{\sigma_c(i)}^c \) represents the predicted constraint parameters, \( CE \) represents the standard cross-entropy function, and \( w_* \) represents the weight balancing term.

\begin{figure}[t]
        \centering
        \includegraphics[width=0.49\textwidth,keepaspectratio]{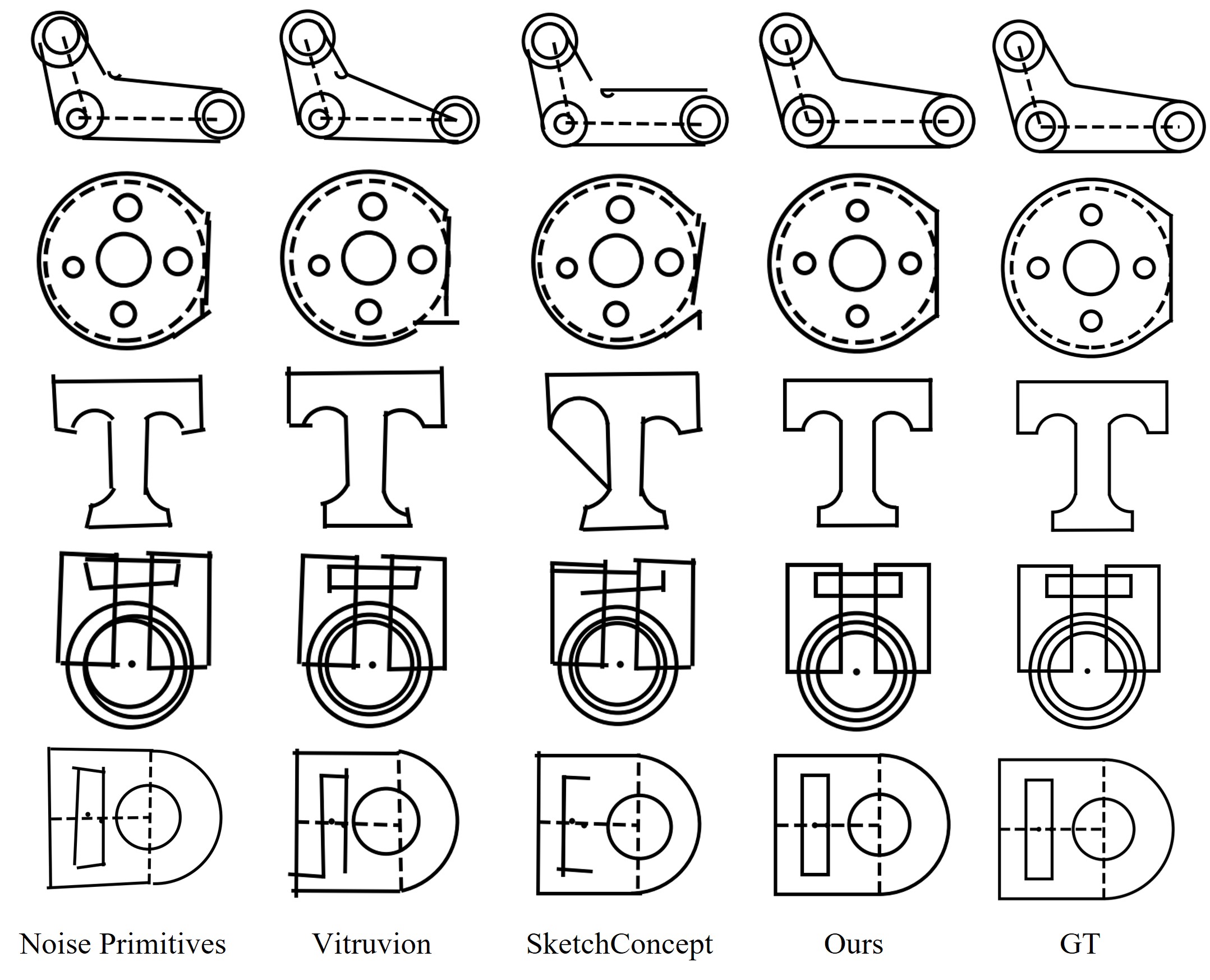}
        \caption{Comparison with the state-of-art constraints parsing methods on the noisy primitives.}
        \label{fig:constraint_image}
\end{figure}

\section{Experiments}
\label{Experiments}
\subsection{Metrics}
To measure the performance of our model, we have proposed specific metrics for different modules and clarified the distinctions between these module metrics. All numbers in \textbf{bold} indicate superior performance.

\textbf{Primitive Metrics.} The metrics for the primitive model, we define three accuracy metrics, namely Primitive Type Accuracy $(ACC^p_{type})$, Boolean Flag Accuracy $(ACC_{flag})$, and Primitive Parameters Accuracy $(ACC^p_{par})$.
\begin{equation}
       ACC^p_{type} =  \frac{1}{K_p}\sum_{i=1}^{K_p}\mathbb{I}[t^p_i=\hat{t}^p_{\sigma_p(i)}],
\end{equation}
\begin{equation}
        ACC_{flag} =  \frac{1}{K_p}\sum_{i=1}^{K_p}\mathbb{I}[f_i=\hat{f}_{\sigma_p(i)}],
\end{equation}
\begin{equation}
ACC^p_{par} = \frac{1}{K_p}\sum_{i=1}^{K_p}\mathbb{I}{[|p^p_i-\hat{p}^p_{\sigma_p(i)}|}\leq\eta],
\label{acc_p_param}
\end{equation}
where \( \sigma_p \) represents the primitive index mapping relationship, $t^p_i$ and $\hat{t}^p_{\sigma_p(i)}$ is the GT and predicted primitive type, $f_i$ and $\hat{f}_{\sigma_p(i)}$ is the GT and predicted boolean flag, 
$p^p_i$ and $\hat{p}^p_{\sigma_p(i)}$ are the GT and predicted primitive parameters and $\mathbb{I}$ is an indicator function that equals 1 when the condition is satisfied and 0 otherwise.

Note that predicted primitive parameters $\hat{p}^p_{\sigma_p(i)}$ and GT primitive parameters $p^p_i$ are both quantized into 6-bit integers. In practice, we adopt the same strategy as DeepCAD \cite{wu2021deepcad}, allowing the error in primitive coordinates to be within a threshold $\eta$ =1 (out of 64 levels) to be considered as a correct prediction. Additionally, the constraint model can also correct the biases generated by the primitive model.

\textbf{Constraint Metrics.} The metrics for the constraint model, we define two accuracy metrics, namely Constraint Type Accuracy $(ACC^c_{type})$ and Constraint Parameters Accuracy $(ACC^c_{par})$,
\begin{equation}
        ACC^c_{type} = \frac{1}{K_c} \sum_{i=1}^{K_c}\mathbb{I}[t^c_i=\hat{t}^c_{\sigma_c(i)}],
\end{equation}
\begin{equation}
        ACC^c_{par} =\frac{1}{K_c} \sum_{i=1}^{K_c}\mathbb{I}[p^c_i=\hat{p}^c_{\sigma_c(i)}],
\label{acc_c_index}
\end{equation}
where \( \sigma_c \) denotes the constraint index mapping relationship, $t^c_i$ and $\hat{t}^c_{\sigma_c(i)}$ are the GT and predicted constraint type, $p^c_i$ and $\hat{p}^c_{\sigma_c(i)}$ are the GT and predicted constraint parameters. Eq. \ref{acc_p_param} are used to calculate parameters' accuracy and {$\mathbb{I}$ is an indicator function that equals 1 when the condition is satisfied and 0 otherwise. However, we do not employ a threshold in Eq. \ref{acc_c_index} because the emphasis is on constraint parameters, which are closely associated with the indices of primitives.
\begin{table}[t]
    \caption{The evaluations of constraint model on the noisy sketch test set of Vitruvion \cite{seff2021vitruvion} dataset.} 
    \label{tab:table_five}
    \resizebox{\linewidth}{!}{
    \begin{tabular}{c|c|c|c|c|c}
        \hline
        \multirow{2}{*}{\begin{tabular}[c]{@{}c@{}}Training\\ Regimen\end{tabular}} & \multirow{2}{*}{Model} & \multicolumn{2}{c}{\begin{tabular}[c]{@{}c@{}}Noiseless Testing\end{tabular}} & \multicolumn{2}{|c}{\begin{tabular}[c]{@{}c@{}}Noisy Testing\end{tabular}} \\ \cline{3-6} 
         &  & \begin{tabular}[c]{@{}c@{}}$ACC^c_{type}$\end{tabular} & \begin{tabular}[c]{@{}c@{}}$ACC^c_{par}$\end{tabular} & \begin{tabular}[c]{@{}c@{}}$ACC^c_{type}$\end{tabular} & \begin{tabular}[c]{@{}c@{}}$ACC^c_{par}$\end{tabular} \\ \hline
        \multirow{3}{*}{\begin{tabular}[c]{@{}c@{}}Noiseless \\ Training\end{tabular}} & Vitruvion & 57.45\% & 47.89\% &56.18\% & 43.73\% \\ \cline{2-6} 
         & SketchConcept &78.53\%  &76.21\%  &71.16\%  &69.78\%  \\ \cline{2-6} 
         & Ours & \textbf{94.29\%} &\textbf{ 97.07\%} & \textbf{83.39\%} & \textbf{91.59\%} \\ \hline
        \multirow{3}{*}{\begin{tabular}[c]{@{}c@{}}Noisy \\ Training\end{tabular}} & Vitruvion & 59.67\% & 49.86\% & 58.23\% & 49.52\%  \\\cline{2-6} 
         & SketchConcept &80.12\%  &81.15\% &78.12\%  &79.88\%   \\ \cline{2-6} 
         & Ours & \textbf{96.46\%} & \textbf{96.94\%} & \textbf{96.39\%} & \textbf{96.83\%} \\ \hline
    \end{tabular}
}
\end{table}
\begin{table}[t]
   \centering
    \caption{The evaluations of the constraint model on the noisy sketch test set of SketchConcept \cite{yang2022discovering} dataset.} 
    \label{tab:table_six} 
    \resizebox{\linewidth}{!}{ 
        \begin{tabular}{c|c|c|c|c|c}
    \hline
    \multirow{2}{*}{\begin{tabular}[c]{@{}c@{}}Training\\ Regimen\end{tabular}} & \multirow{2}{*}{Model} & \multicolumn{2}{c}{\begin{tabular}[c]{@{}c@{}}Noiseless Testing\end{tabular}} & \multicolumn{2}{|c}{\begin{tabular}[c]{@{}c@{}}Noisy Testing\end{tabular}} \\ \cline{3-6} 
     &  & \begin{tabular}[c]{@{}c@{}}$ACC^c_{type}$\end{tabular} & \begin{tabular}[c]{@{}c@{}}$ACC^c_{par}$\end{tabular} & \begin{tabular}[c]{@{}c@{}}$ACC^c_{type}$\end{tabular} & \begin{tabular}[c]{@{}c@{}}$ACC^c_{par}$\end{tabular} \\ \hline
    \multirow{3}{*}{\begin{tabular}[c]{@{}c@{}}Noiseless \\ Training\end{tabular}} & Vitruvion & 60.03\% & 49.11\% & 53.28\% & 45.19\% \\ \cline{2-6} 
     & SketchConcept &77.23\%  &76.79\%  &68.21\%  &71.69\%  \\ \cline{2-6} 
     & Ours & \textbf{94.26\%} & \textbf{97.83\%} & \textbf{81.97\%} & \textbf{92.92\%} \\ \hline
    \multirow{3}{*}{\begin{tabular}[c]{@{}c@{}}Noisy \\ Training\end{tabular}} & Vitruvion &61.38\%  &50.23\%  &57.37\%  &48.22\%  \\ \cline{2-6} 
     & SketchConcept &79.33\%  &78.52\% &77.83\%  &78.13\%   \\ \cline{2-6} 
     & Ours & \textbf{97.04\%} & \textbf{97.29\%} & \textbf{96.99\%} & \textbf{97.25\%} \\ \hline
    \end{tabular}
    }
\end{table}
\begin{table*}[t]
\centering
\begin{minipage}[b]{0.48\linewidth}
    \centering
    \caption{The effects of different patch numbers of primitive model.} 
        \label{ablation_study_patch_primitive_model}
    \resizebox{\linewidth}{!}{ 
    \begin{tabular}{c|c|c|c|c}
    \hline
    \multicolumn{1}{c|}{\begin{tabular}[c]{@{}c@{}}Patch\\ Number\end{tabular}} & \multicolumn{1}{c|}{\begin{tabular}[c]{@{}c@{}}Patch\\ Size\end{tabular}} & $ACC^p_{type}$ & $ACC_{flag}$ & $ACC^p_{par}$ \\ \hline
    \multicolumn{1}{c|}{16} & \multicolumn{1}{c|}{32*32} &90.20\%  &92.15\%  &77.10\%   \\ \hline
    \multicolumn{1}{c|}{32} & \multicolumn{1}{c|}{16*32} &91.50\%  &93.76\%  &82.11\%  \\ \hline
    \multicolumn{1}{c|}{64} & \multicolumn{1}{c|}{16*16} &\textbf{95.11\%}  &\textbf{95.00\%}  &\textbf{87.26\%}   \\ \hline
    \multicolumn{1}{c|}{128} & \multicolumn{1}{c|}{16*8} &94.01\%  &94.58\%  &85.53\%   \\ \hline
    \end{tabular}
    }
\end{minipage}\hfill
\begin{minipage}[b]{0.48\linewidth}
    \centering
    \caption{The effects of different primitive parameter Loss of primitive model.} 
        \label{ablation_study_Loss_primitive_model}
    \resizebox{\linewidth}{!}{ 

    \begin{tabular}{c|c|c|c|c}
    \hline
    \begin{tabular}[c]{@{}c@{}}Parameter Loss\\ Type\end{tabular} & $ACC^p_{type}$ & $ACC_{flag}$ & $ACC^p_{par}$ & $CD$ \\ \hline
    Classification &\textbf{95.11\%}  &\textbf{95.00\%}  &\textbf{87.26\%}  &\textbf{0.020}  \\ \hline
    Regression &89.07\%  &91.10\%  & -- & 0.062  \\ \hline
    \end{tabular}
    }
    
\end{minipage}
\end{table*}

\begin{table*}[t]
\centering
\begin{minipage}[b]{0.48\linewidth}
    \centering
    \caption{The effects of pointer module of constraint model.} 
    \label{ablation_study_Pointer_constraint_model}
    \resizebox{\linewidth}{!}{ 
     \begin{tabular}{c|c|c|c|c|c|c}\hline
    \multirow{2}{*}{\begin{tabular}[c]{@{}c@{}}Parameter\\ Type\end{tabular}} & \multirow{2}{*}{\begin{tabular}[c]{@{}c@{}}Pointer\\ Module\end{tabular}} & \multirow{2}{*}{\begin{tabular}[c]{@{}c@{}}Encoding\\ Method\end{tabular}} & \multicolumn{2}{c|}{\begin{tabular}[c]{@{}c@{}}Noiseless Testing\end{tabular}} & \multicolumn{2}{c}{\begin{tabular}[c]{@{}c@{}}Noisy Testing\end{tabular}} \\ \cline{4-7} 
     &  &  & $ACC^c_{type}$ & $ACC^c_{par}$ & $ACC^c_{type}$ & $ACC^c_{par}$ \\ \hline
    6-bit Integer & $\times$ & Embedding & 94.67\%  &95.13\%  &94.21\%  &95.03\%  \\ \hline
    6-bit Integer & \checkmark & Embedding & \textbf{96.46\%} & \textbf{96.94\%} & \textbf{96.39\%} & \textbf{96.83\%}  \\ \hline
    \multicolumn{3}{c|}{$\triangle$} &+1.79\% & +1.81\% & +2.18\% & +1.80\%  \\ \hline
    \end{tabular}
    }
\end{minipage}\hfill
\begin{minipage}[b]{0.48\linewidth}
    \centering
    \caption{The effects of different encoding methods of constraint model.} 
    \label{ablation_study_encoding_constraint_model}
    \resizebox{\linewidth}{!}{ 
     \begin{tabular}{c|c|c|c|c|c|c}
     \hline
    \multirow{2}{*}{\begin{tabular}[c]{@{}c@{}}Parameter\\ Type\end{tabular}} & \multirow{2}{*}{\begin{tabular}[c]{@{}c@{}}Pointer\\ Module\end{tabular}} & \multirow{2}{*}{\begin{tabular}[c]{@{}c@{}}Encoding\\ Method\end{tabular}} & \multicolumn{2}{c|}{\begin{tabular}[c]{@{}c@{}}Noiseless Testing\end{tabular}} & \multicolumn{2}{c}{\begin{tabular}[c]{@{}c@{}}Noisy Testing\end{tabular}} \\ \cline{4-7} 
     &  &  & $ACC^c_{type}$ & $ACC^c_{par}$ & $ACC^c_{type}$ & $ACC^c_{par}$ \\ \hline
    6-bit Integer & \checkmark & Embedding & \textbf{96.46\%} & \textbf{96.94\%} & \textbf{96.39\%} & \textbf{96.83\%}  \\ \hline
    Float & \checkmark & Sin Cos &89.79\%  &92.86\%  &89.53\%  &92.67\%  \\ \hline
    Float & \checkmark & MLP &93.33\%  &92.46\%  &92.76\%  &91.89\%  \\ \hline
    \end{tabular}
    }
\end{minipage}
\end{table*}

\begin{table*}[t]
\centering
\begin{minipage}[b]{0.48\linewidth}
    \centering
    \caption{The effects of different Transformer layers of primitive model.} 
    \label{Transformer_primitive}
    \begin{tabular}{c|c|c|c}
    \hline
    \begin{tabular}[c]{@{}c@{}}Transformer\\ Layers\end{tabular} & $ACC^p_{type}$ & $ACC_{flag}$ & $ACC^p_{par}$ \\ \hline
    12 &\textbf{95.11\%}  &\textbf{95.00\%}  &\textbf{87.26\%}  \\ \hline
    8 &93.11\%  &94.72\%  &84.34\%  \\ \hline
    4 &92.17\%  &93.53\%  &84.31\%  \\ \hline
    \end{tabular}
\end{minipage}\hfill
\begin{minipage}[b]{0.48\linewidth}
    \centering
    \caption{The effects of different Transformer layers of constraint model.} 
    \label{Transformer_Constraint}
    \begin{tabular}{c|c|c|c|c}\hline
    \multirow{2}{*}{\begin{tabular}[c]{@{}c@{}}Transformer\\ Layers\end{tabular}} & \multicolumn{2}{c|}{\begin{tabular}[c]{@{}c@{}}Noiseless Testing\end{tabular}} & \multicolumn{2}{c}{\begin{tabular}[c]{@{}c@{}}Noisy Testing\end{tabular}} \\ \cline{2-5} 
     & $ACC^c_{type}$ & $ACC^c_{par}$ & $ACC^c_{type}$ & $ACC^c_{par}$ \\ \hline
    12 & \textbf{96.46\%} & \textbf{96.94\%} & \textbf{96.39\%} & \textbf{96.83\%}  \\ \hline
    8 &95.92\%  &95.48\%  &95.82\%  &95.35\%  \\ \hline
    4 &96.04\%  &93.45\%  &95.95\%  &93.33\%  \\ \hline
    \end{tabular}
\end{minipage}
\end{table*}

\begin{table*}[h]
        \centering
    \begin{minipage}{0.48\linewidth}
        \centering
        \caption{The effects of different weights of primitive model.} 
        \label{primitive_ablation}
        \begin{tabular}{c|c|c|c}
        \hline
        \multicolumn{1}{c|}{$[\omega^p_{t},\omega_{f},\omega^p_{p}]$} & \multicolumn{1}{c|}{$ACC^p_{type}$} & \multicolumn{1}{c|}{$ACC_{flag}$} & \multicolumn{1}{c}{$ACC^p_{par}$} \\ \hline
        $[1,1,1]$ &\textbf{96.58\%}  &\textbf{98.57\%}  &85.03\%  \\  \hline
        $[1,1,5]$ &95.11\%  &95.00\%  &87.26\%  \\  \hline
        $[1,1,7]$ &94.80\%  &95.02\%  &\textbf{88.73\%}  \\  \hline
        $[1,1,10]$ &94.08\%  &94.33\%  &87.03\%  \\ \hline
        \end{tabular}
    \end{minipage}\hfill
    \begin{minipage}{0.48\linewidth}
        \centering
        \caption{The effects of different weights of constraint model.} 
        \label{constraint_ablation}
        \resizebox{\linewidth}{!}{ 
        \begin{tabular}{c|c|c|c|c}
        \hline
         & \multicolumn{2}{c|}{\begin{tabular}[c|]{@{}c@{}}Noiseless Testing\end{tabular}} & \multicolumn{2}{c}{\begin{tabular}[c]{@{}c@{}}Noisy Testing\end{tabular}} \\ \hline
        $[\omega^c_{t},\omega^c_{p}]$ & $ACC^c_{type}$ & $ACC^c_{par}$ & $ACC^c_{type}$ & $ACC^c_{par}$ \\ \hline
        $[1,1]$ & \textbf{96.46\%} & 96.94\% & \textbf{96.39\%} & 96.83\% \\  \hline
        $[1,3]$ & 92.49\% & 97.55\% & 92.27\% & 97.32\%  \\  \hline
        $[1,7]$ & 90.42\% & \textbf{97.80\%} & 90.24\% & \textbf{97.77\%}  \\ \hline
        \end{tabular}
        }
    \end{minipage}
\end{table*}

\begin{table*}[h]
        \centering
    \begin{minipage}{0.48\linewidth}
        \centering
        \caption{The effects of constraint network.} 
        \label{Constraint_improve_parameter_acc}
    \begin{tabular}{c|c|c|c}
    \hline
    \begin{tabular}[c]{@{}c@{}}\end{tabular} & $ACC^p_{type}$ & $ACC_{flag}$ & $ACC^p_{par}$ \\ \hline
    Without Constraint &95.11\% &95.00\%  &87.26\%  \\ \hline
    With Constraint &95.11\%  &95.00\%  &\textbf{89.31\%}  \\ 
 \hline
    \end{tabular}
    \end{minipage}\hfill
        \begin{minipage}{0.48\linewidth}
        \centering
        \caption{The quantization and prediction errors.} 
        \label{Error_experiment}
    \begin{tabular}{c|c}
    \hline
    \begin{tabular}[c]{@{}c@{}}\end{tabular} & $CD$\\ \hline 
    Quantization Error &0.0053  \\  \hline
    Prediction Error   &0.0202 \\ 
 \hline
    \end{tabular}
    \end{minipage}
\end{table*}

\subsection{Comparison with state-of-art}
\label{subsec:Comparison}
We conducted a comprehensive evaluation of the primitive model and the constrained model on two datasets, Vitruvion \cite{seff2021vitruvion} dataset and SketchConcept \cite{yang2022discovering} dataset, incorporating both qualitative and quantitative results.

\textbf{Primitive Evaluation.} As shown in Table \ref{tab:table_two_three}, we conducted quantitative comparisons on two datasets, evaluating the performance of the primitive model under noisy sketch training and testing conditions. Simultaneously, in Fig. \ref{fig:primitive_image}, we provided a qualitative representation showcasing the results of our experimental comparisons.

\textbf{Constraint Evaluation.} As shown in Table \ref{tab:table_five} and Table \ref{tab:table_six}, following Vitruvion \cite{seff2021vitruvion} noise primitives method, we conducted a quantitative comparison of the network on two different datasets. We compared different training methods and evaluated them under diverse testing conditions. As shown in Fig. \ref{fig:constraint_image}, we performed a qualitative analysis to further validate the quantitative results.

\subsection{Experimental Analysis.} 
In this section, we analyze the experimental results from the previous section.

\textbf{Primitive Analysis.} As demonstrated in Tables \ref{tab:table_two_three}, our approach exhibits the best performance in the presence of noise. If an autoregressive model of Vitruvion \cite{seff2021vitruvion} is directly used, in the inference phase, the previous $N$-1 steps results are required as the input to infer the N-$th$ step result, which process is repeated, jointly predicting both primitive types and parameters until the end token appears. However, this architecture is not only very time-consuming but also prone to accumulating errors if the output from earlier steps is biased or incorrect. Such biased or incorrect steps can significantly impact all subsequent output results. We approach this challenge by separately predicting the primitive types and parameters rather than treating them together. This separation avoids the complexities arising from this mixture and reflects in the improved performance of metrics $ACC^p_{type}$, $ACC_{flag}$ and $ACC^p_{par}$.

As shown in Fig. \ref{fig:primitive_image}, during the practical testing of Vitruvion \cite{seff2021vitruvion}, a notable issue arose where the end token appeared prematurely, leading to the premature termination of the network's output sequence. This problem directly impacted the visual representation of sketch test results, causing them to appear incomplete, as evident in the case of (Row 2, Column 2). Furthermore, a common issue associated with generative models, the gradual accumulation of errors, further compromised the accuracy of the results, as indicated in (Row 5, Column 2). However, through our approach of separately predicting primitive types and parameters, we managed to overcome these challenges. Even when dealing with complex models, we demonstrated improved effectiveness, as illustrated in (Row 4, Column 3, and Row 5, Column 3).

\textbf{Constraint Analysis.} As demonstrated in Tables \ref{tab:table_five} and \ref{tab:table_six}, our approach exhibits the best performance in the presence of noise. Since the constraint model in Vitruvion \cite{seff2021vitruvion} is built on an autoregressive model, it requires using the output of the previous $N$-1 steps as input to infer the result of the N-$th$ step. This process continues iteratively until an end token appears. Consequently, this constraint model faces similar issues as its primitive model, including the cumbersome iterative processes and the error accumulation. In contrast to the Vitruvion \cite{seff2021vitruvion} method, which combines the output of constraint types and parameters and focuses on constraints applied to primitives or sub-primitives, our method keeps these aspects distinct. Furthermore, the SketchConcept \cite{yang2022discovering} network's predictive output not only includes primitive types but also constraint types without clear separation between the two. Simultaneously, it does not establish a direct binding of constraint parameters to primitives. This could result in constraints being simultaneously applied to primitives and other constraints, making it difficult to directly apply to hand-drawn sketch analysis tasks. However, we solve the above problems by separately predicting the type and parameters of constraints and binding constraint parameters to the indices of primitives using a pointer module.

\begin{figure}[t]
    \centering
    \includegraphics[width=0.49\textwidth,keepaspectratio]{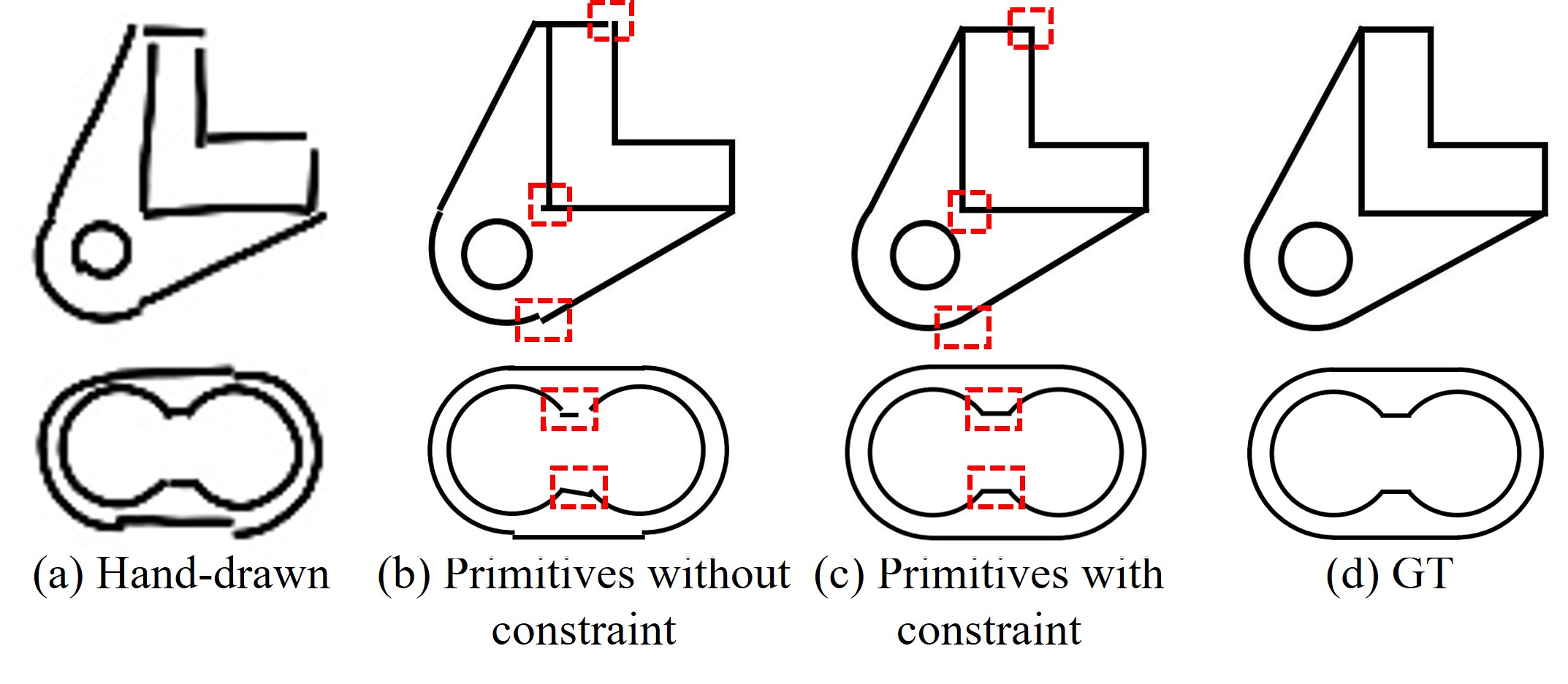}
    \caption{Quantitative analysis of experimental results with and without the constraint network.} 
    \label{fig:Primitive_in_constraint}
\end{figure}
In the examples presented in Fig. \ref{fig:constraint_image}, we can observe variations in the performance of different methods when recovering constraint relationships. Specifically, in the Vitruvion \cite{seff2021vitruvion} approach, the premature appearance of the end token leads to an incomplete capture of constraint relationships for all primitives, causing issues in forming a coherent sketch (see Row 3, Column 2). Similarly, the SketchConcept \cite{yang2022discovering} network makes errors in predicting constraint parameters, as seen in the case of (Row 3, Column 3). This results in problems when applying constraint relationships to the sketch, particularly evident in the incorrect connections between line and arc. However, by introducing the pointer module in our method, we can overcome these challenges. Our approach allows constraint parameters to be directly associated with the indices of primitives, as illustrated in (Row 1, Column 4) of the figure. Even when dealing with complex shapes, our method accurately captures constraint relationships between primitives, facilitating the recovery of a continuous sketch. Compared to the Vitruvion \cite{seff2021vitruvion} and SketchConcept \cite{yang2022discovering} methods, our approach offers the advantage of directly linking constraint parameters to primitive indices.

\subsection{Ablation Studies}
Due to the larger sample size of the Vitruvion \cite{seff2021vitruvion} dataset, we will conduct our ablation experiments using this dataset. Under noisy conditions, we train and test both the primitive model and the constraint model separately, comparing the metric's variations under different conditions. 


\textbf{Patch Number Ablation.}
As depicted in Table \ref{ablation_study_patch_primitive_model}, indiscriminately altering the number of patches does not yield improved metric indicators.

\textbf{Parameter Loss Ablation.}
As shown in Table \ref{ablation_study_Loss_primitive_model}, we have replaced the classification loss for predicting primitive parameters in the primitive model with regression loss. The difference between losses of different types of primitive parameters is measured by calculating the Chamfer Distance ($CD$).

\textbf{Pointer Module Ablation.}
As shown in Table \ref{ablation_study_Pointer_constraint_model}, the consideration of utilizing the pointer module is contingent upon its necessity; if not used, it is to be replaced with a final MLP-based classification network with $K_p$, where \( K_p \) is the number of primitives contained in the ground truth sketch. The performance improvement brought about by using the pointer module is shown in the last row of the table. We found that using the pointer module not only performs better on metrics but also enhances the interpretability of the network.

\textbf{Encoding Methods Ablation.}
As shown in Table \ref{ablation_study_encoding_constraint_model}, we explored the scenario where the primitive parameters are expressed in different forms. Specifically, we investigated the conversion of floating-point numbers to 6-bit integers and their unaltered representation. We compared the impacts of these different conditions on the constraint network. In this experiment, we used traditional positional encoding (using Sin and Cos function) and MLP (directly learning to map primitive parameters to high-dimensional vectors) for comparison.

\textbf{Transformer Layers Ablation.}
As shown in Table \ref{Transformer_primitive} and Table \ref{Transformer_Constraint}, the reduction of Transformer layers in both the primitive model and the constraint model leads to a varying degree of decrease in all performance metrics.

\textbf{Loss Weight Ablation.}
Under the noise condition, the primitive model and the constraint model are trained and tested respectively, and the performance changes under different loss weights are compared. The effects of different weights of the primitive model and the constraint model are shown in tables \ref{primitive_ablation} and \ref{constraint_ablation}.
In this context, $\omega_{*}$ in Tables \ref{primitive_ablation} and \ref{constraint_ablation} represents the weight balancing all loss terms.

\textbf{The effects of constraint network.} In Table \ref{Constraint_improve_parameter_acc} and Fig. \ref{fig:Primitive_in_constraint}, we show the prediction results with and without the constraint network, respectively. Table \ref{Constraint_improve_parameter_acc} presents the improvement in the accuracy of primitive parameters, $ACC^p_{par}$, achieved by the constraint network, while the accuracies of primitive types, $ACC^p_{type}$, and boolean flags, $ACC_{flag}$, remain unchanged. This is mainly because the constraint relationships only adjust the primitive parameters. Fig. \ref{fig:Primitive_in_constraint} (b) shows the prediction results of the primitive network, while Fig. \ref{fig:Primitive_in_constraint} (c) shows the results after applying the constraint relationships to correct the primitive parameters.

\textbf{The quantization and prediction errors.} As shown in Table \ref{Error_experiment}, we further measure the quantization error introduced by integerizing floating-point primitive parameters to 6-bit integers in CAD sketches. Additionally, we evaluate the prediction errors between the model predictions and the real CAD sketches.



\section{Conclusion}
\label{Conclusion}

We convert the sketch analysis task into a Set Prediction problem, accurately extracting primitives and constraints from the given input. Our method exhibits superior performance on two large-scale publicly available datasets. Additionally, our approach is straightforward to implement and has a flexible architecture that can extend to free-form curves.
In contrast, it avoids the iterative time consumption and error accumulation issues faced by Vitruvion \cite{seff2021vitruvion}, as well as the computational overhead associated with maintaining the program library in SketchConcept \cite{yang2022discovering}.

\textbf{Limitation and Future Work.} We divide the sketch analysis task into two sub-tasks: primitive parsing and constraint parsing, which are independent of each other. In future work, we plan to design an end-to-end neural framework so that the model takes only hand-drawn sketches as input and outputs both primitives and constraints. Additionally, current large-scale CAD sketch datasets (e.g.,  Vitruvion \cite{seff2021vitruvion}) focus on the four primitive types (line, arc, cycle, point), which lack annotation data for free-form curves. The lack of training samples can also lead to limitations in algorithm design and validation. Therefore, the construction of a high-quality CAD sketch dataset containing a large number of free-form curves will also be a focal point of future work.
\bibliography{main} 
\bibliographystyle{plain} 

\end{document}